\documentclass{article}
\usepackage{spconf,amsmath,graphicx,booktabs,epstopdf}

\title{Weakly supervised instance segmentation using hybrid networks}
\name{\begin{tabular}{c}
     Shisha Liao$^{\star}$ \qquad Yongqing Sun$^{\dagger}$ \qquad Chenqiang Gao$^{\star}$ \\ \qquad Pranav Shenoy K P$^{\ddagger}$ \qquad Song Mu$^{\star}$  \qquad Jun Shimamura$^{\dagger}$ \qquad Atsushi Sagata$^{\dagger}$
\end{tabular}}

\address{ \begin{tabular}{c}
$\star$ School of Communication and Information Engineering, \\ Chongqing University of Posts and Telecommunications, Chongqing, China  \qquad \\
$\dagger$ NTT Media Intelligence Laboratories, Japan  \qquad \\ $\ddagger$ Georgia Institute of Technology, GA, USA
\end{tabular}}

\begin{document}
%
\maketitle
\begin{abstract}
Weakly-supervised instance segmentation, which could greatly save labor and time cost of pixel mask annotation, has attracted increasing attention in recent years.
The commonly used pipeline firstly utilizes conventional image segmentation methods to automatically generate initial masks and then use them to train an off-the-shelf segmentation network in an iterative way.
However, the initial generated masks usually contains a notable proportion of invalid masks which are mainly caused by small object instances.
Directly using these initial masks to train segmentation models is harmful for the performance.
To address this problem, we propose a kind of hybrid networks in this paper.
In our architecture, there is a principle segmentation network which is used to handle the normal samples with valid generated masks.
In addition, a complementary branch is added to handle the small and dim objects without valid masks.
Experimental results indicate that our method can achieve significantly performance improvement both on the small object instances and large ones, and outperforms all state-of-the-art methods.

\end{abstract}
\section{Introduction}
\label{sec:intro}
Instance segmentation, which serves as a fundamental task for a broad set of vision applications, such as remote sensing cite{remotesensing}, medical imaging \cite{medicalimaging}, and automatic drive \cite{driving}, has attracted extensive attention and made large progress in the recent years.
Most state-of-the-art methods \cite{maskrcnn, Panet} rely on large-scale dense annotations for training deep networks and show promising performances among the challenging benchmark datasets, including COCO \cite{lin2014microsoft}, CityScapes \cite{CityScapes} and PASCAL VOC \cite{everingham2010pascal}.
However, annotating pixel-level labels for object instances is particularly expensive and time-consuming \cite{russakovsky2015imagenet}.
Comparing with complex and enormous pixel-level masks, some weakly annotations are much easier to obtain, e.g., points, scribbles, bounding boxes and image-level labels.
Therefore, investigating the potentials of weakly supervised instance segmentation can effectively mitigate the labor cost, showing great practical significance.

\begin{figure}[t]
\centering\includegraphics[width=8cm]{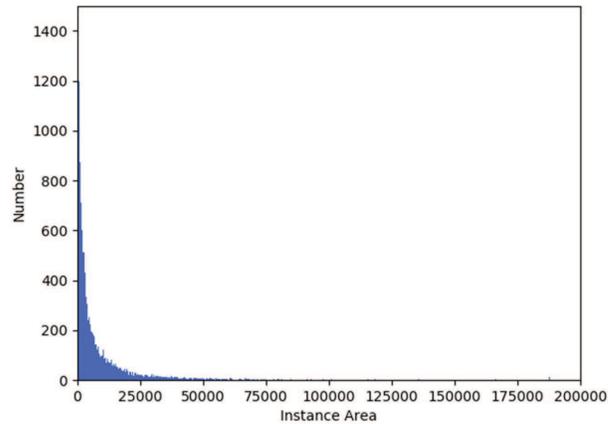}
\caption{The distribution of different sizes (areas) within initial invalid object instances from GrabCut \cite{GrabCut}. The invalid masks concentrate on small object instances}
\label{fig:distribution}
\end{figure}

In the weakly supervised instance segmentation realm, bounding box annotations are widely utilized due to two aspects.
On one hand, bounding boxes provide precise position and category information. On the other hand, they can be used as a prior information for conventional methods, e.g., GrabCut \cite{GrabCut} and MCG \cite{MCG}, to generate initial pixel-level mask labels, abbreviated as \emph{mask} in the follows. BoxSup \cite{Boxsup} generated initial masks using MCG based on bounding boxes and then proposed an iterative training procedure to obtained a good instance segmentation model.
Khoreva et al. \cite{Simple} employed GrabCut and MCG to generate delicate fake masks from the given box-level annotations and then adopted off-the-shelf segmentation network to implement weakly-supervised instance segmentation.
In \cite{Panoptic}, the masks came from the intersection of the labels generated by GrabCut and MCG.
Like BoxSup \cite{Boxsup}, the generated masks were refined in an iterative training fashion.

Among aforementioned methods, the quality of initial segmentation masks from GrabCut, as well as MCG, relies on the sizes of bounding boxes, namely the sizes of object instances.
The masks from large object instances routinely are of good quality, while small object instances tend to be of poor quality. Because we have no ground-truth for the object instance mask, the quality is roughly estimated by the intersection-over-union (IoU) between the bounding box of the automatically generated object instance mask and the ground-truth bounding box.
In this paper, one generated object instance mask is considered to be invalid if the IoU is under 0.5.
Figure \ref{fig:distribution} shows the distribution of object instances in invalid masks obtained by GrabCut.
It can be obviously observed that the scales of objects with invalid masks are of a wide range, but mainly concentrates on small object instances.
Statistically, invalid small object instances whose size is less than 64$\times$64 pixels, accounts for 55.54\% in invalid object instances, but accounts for 76.48\% in all small object instances with both invalid and valid ones.
In contrast, invalid large object instances whose size is bigger than 64$\times$64 pixels only accounts for 23.20\% in all large instances with both invalid and valid ones.
The total invalid object instances within the initial masks can reach 30\% of all masks.
This notable proportion of invalid masks is harmful for directly training models, even using the iterative training technique as done by some of above methods.

Based on above statistical analysis and observations, we propose a hybrid instance segmentation network.
In this architecture, a principle segmentation network is trained using only the samples with valid masks.
Noteworthy, according to our statistics, the majorities of valid samples are large object instances.
Thus, in this network, the training samples are mostly unified and pure, which is beneficial to the overall training.
In addition, an Enhanced-FPN architecture is added to this branch to reduce the transfer distance of low-level feature, providing more localization information.
For the invalid object instance masks which have correct bounding boxes, we design a complementary branch to handle these hard samples which mainly consist of small and dim object instances as discussed before.
The proposed architecture is evaluated on the validation set of PASCAL VOC 2012, and the experimental results reveal that our method can achieve significant improvement on the aforementioned difficult samples, showing the effectiveness of the complementary framework.

\section{THE PROPOSED METHOD}
\label{sec:method}
\begin{figure*}[t]
\centering
\includegraphics[scale=0.7]{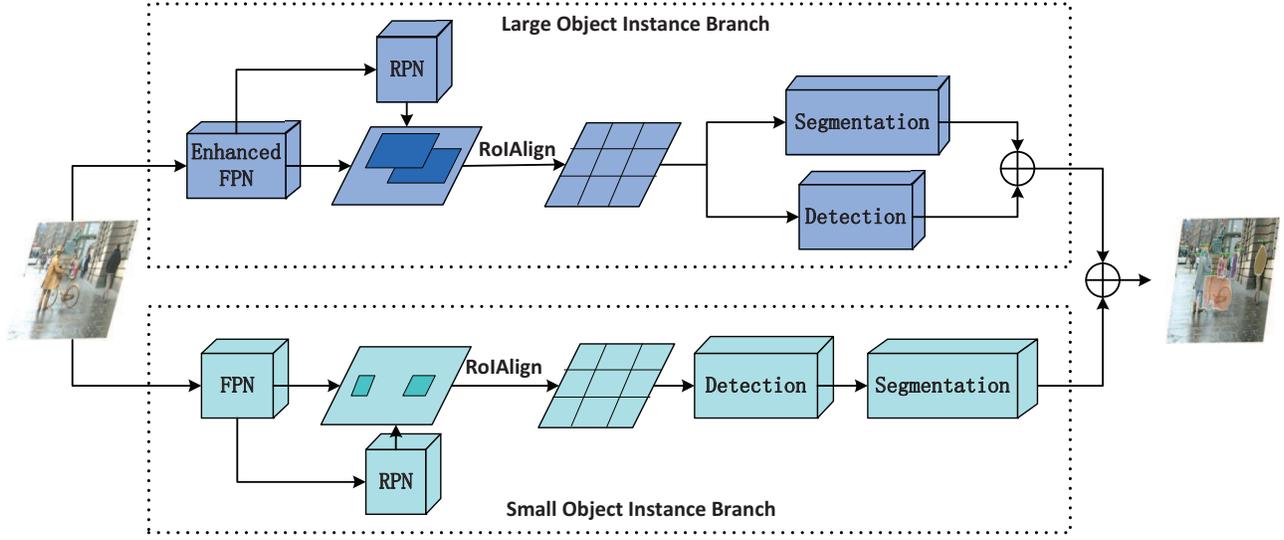}
\caption{Illustration of our framework.}
\label{fig:franmework}
\end{figure*}

As illustrated in Figure \ref{fig:franmework}, the framework of our method contains two branches: a large object instance branch and a small object instance branch.
The large object instance branch cooperates detection and segmentation simultaneously to handle the large object instance segmentation, while the small object instance branch sequentially conducts detection and segmentation to avoid the omission of small object instances.

In the training stage, we firstly use GrabCut \cite{GrabCut} to automatically obtain initial object instance masks based on given bounding box annotations.
Then, we divide the initial masks into two groups according to the IoU as described previously: valid masks and invalid masks.
The former is used to train the large object instance branch, while the latter is used to train the small object instance branch.
In the test stage, images are simultaneously fed into two branches and the segmentation results from both branches are fused to obtain the final results.
Specifically, all the object instance segmentation results with size less than 64$\times$64 pixels come from the small object instance branch, and all the object instance segmentation results with size more than 64$\times$64 pixels come from the big object instance branch.

\subsection{Large object instance branch}
The large object instance branch is built on Mask R-CNN and improved based on our weakly-supervised task. It is composed of four components, including raw feature extraction (Enhanced-FPN module), proposal generation (RPN  module), bounding-box recognition (Detection module) and mask prediction (Segmentation module).
First, We adopt ResNet-50 with Enhanced-FPN as the backbone.
Specifically, the conventional Feature Pyramid Network (FPN) \cite{FPN} architecture is replaced by our Enhanced-FPN to improve the performance.
The Enhanced-FPN  and implementation details will be introduced in the section \ref{sec:Enhance_FPN}.
Then, RPN \cite{RPN} is utilized to generate proposals.
After handled by the RoIAlign \cite{maskrcnn} operation, each proposal becomes a fixed-size feature map.
Finally, bounding-box recognition and mask prediction are implemented simultaneously through this feature map.
The detection branch conforms to the spirit of the fast R-CNN \cite{fastrcnn} to realize localization and classification, and the segmentation branch uses FCN \cite{FCN} to achieve pixel-level prediction.

\subsection{Small object instance branch}
Small object instance branch served as an important complementary module of large objects branch, focuses on small object instances segmentation.
Without the existence of noisy labels, this branch can get better detection performance.
In addition, the morphological characteristic ensures better segmentation performance of small object instances.
Under the supervision of the box-level annotations, this branch first does object detection by Faster R-CNN \cite{RPN}.
The detection results offer the bounding boxes information for GrabCut to obtain final segmentation. For these segmentation results with poor quality, we replace them with ellipses.
In this process, these basic modules are similar to that of the large object branch.
The main difference between these two branches is the execution fashion of detection module and segmentation modules.
The large object instance branch conducts detection and segmentation simultaneously, whereas the small object instance branch conducts them sequentially.

\subsection{Enhanced-FPN}
\label{sec:Enhance_FPN}

\begin{figure}[t]
    \centering\includegraphics[width=7.5cm]{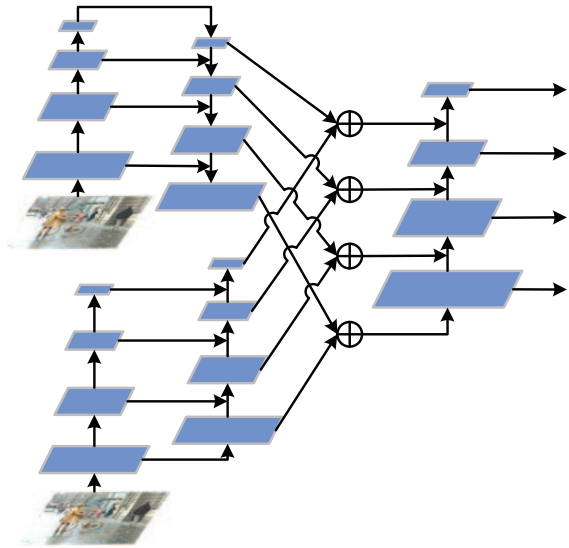}
    \caption{Illustration of our Enhanced-FPN structure.}
    \label{fig:Enhanced-FPN}
    \end{figure}

    \begin{figure}[h]
    \centering\includegraphics[width=8cm]{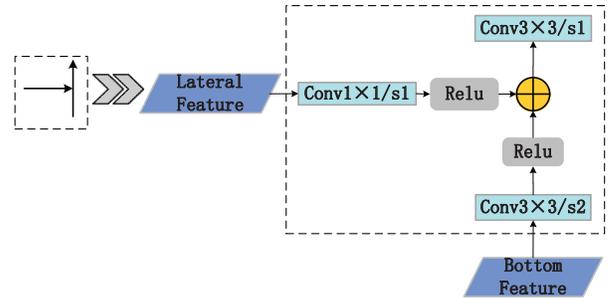}
    \caption{Illustration the connection type of our Enhanced-FPN. (Note: conv $m \times m/sn$ means a convolution whose size is $m \times m$ and whose stride is $n$. Relu means a rectified linear unit)}
    \label{fig:connection}
    \end{figure}
    Enhanced-FPN is improved based on FPN.
    There exists two problems in conventional FPN.
    On the one hand, FPN treats different feature maps unfairly.
    It is well-known that FPN adopts the top-down fusion pattern to increase localization accuracy.
    High-level semantic information is gradually transferred to the low-level feature map, so each low-level feature map includes the information of high-level feature map.
    However, the high-level feature is not enhanced adequately.
    On the other hand, there is a long path from low-level to topmost features and this introduces difficulty to access accurate localization information \cite{Panet}.
    Therefore, we propose an Enhance-FPN to balance the enhancement function of every feature maps, and shorten the propagation distance of bottom feature maps.
    The framework of Enhanced-FPN is illustrated in Figure \ref{fig:Enhanced-FPN}.
    There are multiple structures that the lateral feature map connects with bottom feature map.
    The specific connection type is shown in Figure \ref{fig:connection}.

\subsection{Implementation Details}
\label{ssec:implementation details}
Most hyper-parameters in Mask R-CNN are applied to our first branch. Specifically, We train on 3 GPUs (so effective minibatch size is 6) for 30k iterations, with a learning rate of 0.005 which is decreased by 10 at the 20k and 26K iteration, respectively. We use a weight decay of 0.0002. In addition, we use images with shorter edge randomly sampled from {600, 800} for training and with shorter edge 600 for inference. The longer edge of the images is 1000 for both training and inference. For the second branch, it shares the identical hyper-parameters with the first branch except the learning rate. Its learning rate is 0.0075.

\section{Experiments}
\label{sec:experimrnts}

\subsection{Dataset and metrics}
\label{ssec:dataset and Metrics}
The PASCAL VOC dataset involves 20 semantic categories of objects, which is extensively used in the field of weakly-supervised tasks.
Following the previous work \cite{Simple, PRM}, we utilize additional images from the SBD dataset \cite{SBD} to obtain a training set of 10582 images, and report all of the experimental results on the validation set, including 1449 images.
We adopt the widely used metrics in instance segmentation community, including ${\rm{mAP}}_{0.5}^{\rm{r}}$ and ${\rm{mAP}}_{0.75}^{\rm{r}}$. And the Average Best Overlap (ABO) \cite{ABO} metric is also employed for evaluation to give a different perspective.

\label{ssec: comparision}
\begin{table}[h]
\centering
\caption{Results of different methods on the PASCAL VOC 2012 val.}
\label{tab:the-state-of-the-art}
\setlength{\tabcolsep}{1mm}{
\begin{tabular}{ccccc}
\toprule
Supervision &   Methods & ${\rm{mAP}}_{0.5}^{\rm{r}}$ &  ${\rm{mAP}}_{0.75}^{\rm{r}}$ &  ABO \\
 \midrule
image-level &   SPN \cite{SPN}&          12.7 &       4.4 &       27.1 \\
image-level &   PRM \cite{PRM}&          26.8 &         9 &       37.6 \\
 box-level &   DeepMask\cite{Simple} &   39.4 &        8.1 &       45.8 \\
 box-level & DeepLabBOX\cite{Simple}&    44.8 &       16.3 &       49.1 \\
 box-level &   Ours &                    51.3 &       22.4 &       51.9 \\
\bottomrule
\end{tabular}}
\end{table}

\subsection{Comparison with state-of-the-art methods}

Four state-of-the art methods are selected for comparison, including DeepMask \cite{Simple} and DeepLabBOX \cite{Simple}, soft proposal networks (SPN) \cite{SPN} and peak response maps (PRM) \cite{PRM}.
The former two methods are based on bounding-box level with the same configuration for weak supervised instance segmentation, while the latter two ones are based on the image-level label.
The results of different methods are shown in Table \ref{tab:the-state-of-the-art}.
It can be observed that our method obviously outperforms all state-of-the-art methods in terms of all metrics.
We can also see that the bounding-box-based methods are totally better than the image-label-based methods.
This is because bounding boxes offer more precise information than image labels for instance segmentation.
Among three bounding-box-based methods, the performance of our method is evident, especially in terms of metrics of ${\rm{mAP}}_{0.5}^{\rm{r}}$ and  ${\rm{mAP}}_{0.75}^{\rm{r}}$.
This is due to that our method not only improves the instance segmentation of large objects, but also improves the instance segmentation of small objects.
In addition, the performance of our weakly supervised method with only bounding box information is close to our supervised version with precise pixel mask information for training.

\subsection{Quality analysis}
\label{sec:quality analysis}

\begin{figure}[t]
    \setlength{\belowcaptionskip}{-0.3cm}
    \centering
    \includegraphics[width=8cm]{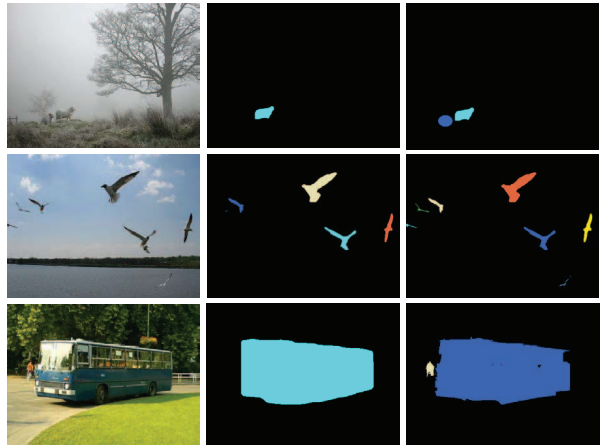}
    \caption {Visual results on PASCAL VOC validation set. The first column is the original images, the second column are the results of the single branch method, and the last column is the results of our method.}
    \label{fig:visual}
\end{figure}

Figure \ref{fig:visual} shows some representative results of our methods.
Note that the second column are results of our large object instance branch trained by all initial generated masks including both invalid and valid masks.
This is similar to commonly used weakly supervised instance segmentation pipeline.
From Figure \ref{fig:visual}, we can observe that our method achieves significant improvement on the small object instance segmentation through adding a complementary small object instance branch.

%
%
%

\section{CONCLUSION}
\label{sec:conclusion}
We propose a novel hybrid segmentation network to handle the invalid mask problem in initial generated masks in the weakly supervised instance segmentation task.
The proposed hybrid network consists of two branches.
One branch cooperates detection and segmentation simultaneously to handle the large object instance segmentation, while the other sequentially conducts detection and segmentation to avoid the omission of small object instances.
Experimental results reveal that our method outperforms state-of-the-art methods, and has obvious advantage on the small object instance segmentation.

\vfill\pagebreak


\bibliographystyle{plain} 
\bibliography{strings,refs}
\end{document}